\documentclass[times,twocolumn,final]{elsarticle}
\usepackage{prletters}
\usepackage{framed,multirow}

\usepackage{amssymb}
\usepackage{latexsym}

\usepackage{url}
\usepackage{xcolor}
\definecolor{newcolor}{rgb}{.8,.349,.1}

\usepackage{graphicx} 
\usepackage{epstopdf} 
\usepackage{hyperref}
\usepackage{tabularx} 
\usepackage{multirow} 
\usepackage{amsthm,amsmath,amssymb,bm,amssymb,mathtools} 
\usepackage{cleveref} 
\DeclarePairedDelimiter{\norm}{\lVert}{\rVert} 

\journal{Pattern Recognition Letters}

\begin{document}

\thispagestyle{empty}
                                                             
\clearpage
\thispagestyle{empty}
\ifpreprint
  \vspace*{-1pc}
\fi

\clearpage
\thispagestyle{empty}

\ifpreprint
  \vspace*{-1pc}
\else
\fi

\ifpreprint
  \setcounter{page}{1}
\else
  \setcounter{page}{1}
\fi

\begin{frontmatter}

\title{An information-theoretic learning model based on importance sampling}
\author[1]{Jiangshe Zhang\corref{cor1}}
\ead{jszhang@mail.xjtu.edu.cn}
\cortext[cor1]{Corresponding author. 
The paper is under consideration at Pattern Recognition Letters.} 
\author[1]{Lizhen Ji}
\author[1]{Fei Gao}
\author[1]{Mengyao Li}

\address[1]{School of Mathematics and Statistics, 
Xi’an Jiaotong University, No.28, Xianning West Road, Xi’an, Shannxi, 710049, P.R. China.}


\begin{abstract}
A crucial assumption underlying the most current theory of machine learning is that 
the training distribution is identical to the test distribution.
However, this assumption may not hold in some real-world applications.
In this paper, we develop a learning model based on principles of information theory 
by minimizing the worst-case loss at prescribed levels of uncertainty.
We reformulate the empirical estimation of the risk functional and the distribution deviation constraint
based on the importance sampling method. 
The objective of the proposed approach is to minimize the loss under maximum degradation
and hence the resulting problem is a minimax problem which
can be converted to an unconstrained minimum problem using the Lagrange method
with the Lagrange multiplier $T$. 
We reveal that the minimization of the objective function under logarithmic transformation
is equivalent to the minimization of the p-norm loss with $p=\frac{1}{T}$.
We applied the proposed model to the face verification task on Racial Faces in the Wild datasets
and showed that the proposed model performs better under large distribution deviations. 
\end{abstract}


\end{frontmatter}

\section{Introduction}
A crucial assumption underlying the most current theory of machine learning is that 
the distribution of the training samples is identical to the distribution of the test samples.
However, it is often violated in practice where the distribution of test data 
deviates from the distribution of training data, 
therefore we need to develop models that work well under realistic violations of this assumption.
In this paper, we assume that the density distribution of future data, instead of being completely unknown, 
is restricted to a class of distributions
and develop a method based on importance sampling and minimax principle\cite{farnia2016minimax}
Although there is considerable freedom in quantizing 
distribution deviation\cite{pardo2018statistical,kailath1967divergence,renyi1961measures},
Kullback–Leibler divergence\cite{kullback1951information,williams1980bayesian}
has been widely used as constraints for regularization in 
bounded rationality \cite{genewein2015bounded,ortega2013thermodynamics}, 
policy optimization\cite{schulman2015trust,hihn2019information} and other learning problems. 
In this work, we use Kullback-Leibler divergence to measure distribution deviation. 

The proposed model aims to minimize the empirical risk in maximum degradation for a given deviation level
and hence the corresponding optimization problem is a minimax problem.
Inspired from the importance sampling method\cite{glynn1989importance,shi2009hierarchical,shi2009neural},
we convert the constraint between the training distribution and test distribution
to a constraint over the importance sampling weights.
Then, we use the Lagrange method to reformulate the constrained minimax optimization problem into 
unconstrained minimum optimization problem which can be solved by available algorithms
like SGD, Adam, etc. 
The innovation of our proposed model is that the future uncertainty is controlled by $T$, 
the Lagrange multiplier in the reformulated optimization problem. 
The objective function of the unconstrained optimization problem is called the Importance Sampling loss (ISloss).
A logarithmic transformation of ISloss is called the Logarithmic Importance Sampling loss (LogISloss).
We reveal that the minimization of LogISLoss is equivalent to the minimization of p-norm with $p=\frac{1}{T}$.

Our proposed importance sampling loss can be applied to many machine learning problems over different applications, 
including regression, classification, clustering, etc. 
In this paper, we adopt the proposed model for the face verification task.  
Face verification\cite{chen2021ring} is a task used to 
determine whether a pair of images belongs to the same individual. 
In the training stage, the model learns a deep feature embedding for an image through the cross entropy loss. 
In the test stgae, the model calculates a cosine similarity
between the two feature embeddings for a given pair of images.
We conducted experiments on Racial Faces in the Wild
datasets(RFW)\cite{Wang_2019_ICCV, MetaRFW} and had the following findings. 
First, the importance sampling weights highlighted the hard classes in the training stage. 
Second, the model trained under LogISloss attained better performance when there was a large distribution deviation.
Third, the model trained under LogISloss emphasized the hard pairs quantized by cosine similarity.

\textbf{Outline of the paper}
\Cref{sec:IS-model} describes our proposed data variation robust learning model
based on importance sampling and the algorithm to solve it. 
The relationship between LogISloss and p-norm is also revealed in this section. 
\Cref{sec:face-verification} applies the proposed model to the face verification task.
\Cref{sec:experiments} shows the experiment results on RFW datasets 
with respect to distribution deviation and hard samples. 
Finally, we conclude this paper in \Cref{sec:conclusion}.

\section{Proposed Method} \label{sec:IS-model}
In the learning problem considered in this paper, we have training data  
$x_1,x_2,\dots, x_N$ which are generated i.i.d according to some (usually unknown) probability density function $q(x)$
and a set of loss function $L(x,\omega)$ and no future data is available for the learning system. 
Our aim is to construct a learning algorithm for a population with an unknown distribution $p(x)$.
We further assume that if $p(x)$ are, instead of being completely unknown,
restricted to a class of distributions, i.e.
\begin{equation}
  \Gamma=\{p(x):KL(p(x)\parallel q(x))=\int p(x)log \left(\frac{p(x)}{q(x)}\right)dx\leq C\}.
\end{equation}
Therefore, our goal becomes to minimize the worst-case expected loss over $\Gamma$. 
A \textit{minimax} approach is applied through minimizing the worst-case loss restricted to this constraint.
\Cref{sec:IS-principle} gives out the principled deviation of the minimax approach
and solves the corresponding optimization problem.
\Cref{sec:log-ISLoss} presents the model under the logarithmic transformation of $L(x,\omega)$
and reveals its relationship with p-norm. 

\subsection{Principle of ISloss}  \label{sec:IS-principle}
The performance of a learning system for a given distribution $p(x)$ is measured by the 
following the risk functional
\begin{equation}
R(\omega)=\int_x L(x,\omega)p(x)dx \label{eq:risk-functional}.
\end{equation}
First, since we don't know exactly the future data distribution $p(x)$ in $\Gamma$,
we need to find a $p(x)$ that maximizes the objective function. 
Given the worst-case distribution $p(x)$,
we aim to find the $\omega$ which minimizes $R(\omega)$.
Therefore, the corresponding optimization problem becomes a minimax problem
\begin{align}
  \begin{split} \label{eq:original-problem}
  \min_{\omega} \max_{p(x)} \quad & R(\omega) =\int_x L(x,\omega) p(x) dx \\
  \text{s.t.}               \quad & KL(p(x)||q(x)) \leq C. \\
  \end{split}
\end{align}
Second, we reformulate the risk functional and the distribution deviation constraint
using the idea of importance sampling method 
where a mathematical expectation with respect to $p(x)$ is approximated by a weighted average of random draws 
from another distribution $q(x)$. 
For any probability density $q(x)$ with $q(x)>0$ whenever $p(x)>0$, 
the risk functional is
\begin{equation}
  R(\omega)=\int_x L(x,\omega)p(x)dx
           =\int_x L(x,\omega){{p(x)\over q(x)} \over {\int_x {p(x)\over q(x)} q(x) dx} }q(x)dx, \label{eq:R-IS} 
\end{equation}
and the KL-divergence between $p(x)$ and $q(x)$ is
\begin{equation}
\begin{split} \label{eq:KL-IS}
KL(p(x)\parallel q(x)) & =\int_x p(x)log \left({p(x)\over q(x)}\right) dx\\
                       & =\int_x {{p(x)\over q(x)} \over {\int_x {p(x)\over q(x)} q(x) dx}} 
                log\left({{p(x)\over q(x)} \over {\int_x {p(x)\over q(x)} q(x) dx} }\right)q(x)dx. 
\end{split}
\end{equation}
Third, we derive the empirical estimation of \eqref{eq:R-IS} and \eqref{eq:KL-IS}.
A standard approach to train the models in statistical learning 
is to use Empirical Risk Minimization(ERM)\cite{vapnik1999nature,farnia2016minimax}. 
ERM learns the prediction rule by minimizing an approximated loss 
under the empirical distribution of samples, which is defined as
\begin{equation}
R_E={1\over N} \sum_{i=1}^{N} L(x_i,\omega) \label{eq:empirical-risk}
\end{equation}
where the subscript E represents \textit{empirical}.
Suppose the observed dataset $D=\{x_1, x_2, \dots, x_N\}$ is $N$ i.i.d samples drawn from $q(x)$.
The self-normalized importance sampling weight\cite{rubinstein2016simulation} for data point $x_i$ is
\begin{equation}
w_i = w(x_i) =  \frac{ \frac{p(x_i)}{q(x_i)} }{\frac{1}{N}\sum_{j=1}^{N} \frac{p(x_j)}{q(x_j)}} \label{eq:is-weights}
\end{equation}
where $\sum_{i=1}^{N} w_i=1$.
$W=[w_i]_{N\times 1}$ is called the importance sampling weight. 
Therefore, the empirical estimation of \eqref{eq:R-IS} is
\begin{equation}
  R_E(\omega) = \frac{1}{N} \sum_{i=1}^{N} \frac{\frac{p(x_i)}{q(x_i)}}
  {\frac{1}{N} \sum_{j=1}^{N} \frac{p(x_j)}{q(x_j)}}L(x_i, \omega) 
  = \sum_{i=1}^{N} L(x_i, \omega) w_i \label{eq:R-IS-proof} 
\end{equation}
and the empirical estimation of \eqref{eq:KL-IS} is
\begin{align}
  \label{eqn:KLpq-proof}
  \begin{split}
  KL_E(p(x) \parallel q(x)) & = \frac{1}{N} \sum_{i=1}^{N} \frac{ \frac{p(x_i)}{q(x_i)} }{\frac{1}{N}\sum_{j=1}^{N} \frac{p(x_j)}{q(x_j)}} 
     log \frac{ \frac{p(x_i)}{q(x_i)} }{\frac{1}{N}\sum_{j=1}^{N} \frac{p(x_j)}{q(x_j)}} \\
  & = \sum_{i=1}^{N} \frac{ \frac{p(x_i)}{q(x_i)} }{\sum_{j=1}^{N} \frac{p(x_j)}{q(x_j)}}
      \left( log\big( \frac{ \frac{p(x_i)}{q(x_i)} }{\sum_{j=1}^{N} \frac{p(x_j)}{q(x_j)}} \big) - log \frac{1}{N} \right) \\
  & = \sum_{i=1}^{N} w_i log  \frac{w_i}{\frac{1}{N}}  = KL(w_i \parallel \{ \frac{1}{N} \}) \\
  & = \sum_{i=1}^{N} w_i log w_i + log N \\
  \end{split}
\end{align}
where $\{ \frac{1}{N}\}$ denotes the discrete uniform distribution with $N$ samples. 
Plugging \eqref{eq:R-IS-proof}  and \eqref{eqn:KLpq-proof} back into \eqref{eq:original-problem},
the optimization problem  \eqref{eq:original-problem} becomes
\begin{align}
  \min_{\omega}\max_{W}  \quad & R_E(\omega,W) =\sum_{i=1}^{N} L(x_i,\omega)w_i, \label{eq:empirical-problem}\\
        \text{s. t.}     \quad & \sum_{i=1}^{N}  w_i log w_i + log(N) \leq C, \label{eq:wi-KL-constraint} \\
                               & \sum_{i=1}^{N}  w_{i}=1, w_{i}\in [0,1], \quad 1\leq i \leq N. \label{eq:wi-constraint}
\end{align}
where \eqref{eq:wi-constraint} is the constraint for the importance sampling weight $W$.
Next, the constrained optimization problem in \eqref{eq:empirical-problem}
can be reformulated to the unconstrained optimization problem using the Lagrange method.
Let the Lagrange multiplier for \eqref{eq:wi-KL-constraint} and \eqref{eq:wi-constraint}
be $T$ and $\lambda$ respectively.
The max problem in \eqref{eq:empirical-problem} can be reformulated as
\begin{equation}
  F_{0}(\omega,W)=\sum_{i=1}^{N} L(x_i,\omega)w_i-T\sum_{i=1}^{N} w_i log w_i-Tlog(N) -\lambda(\sum_{i=1}^{N} w_i-1).
\end{equation}
Here $T>0$ is the temperature that governs the level of randomness of $W$,
which implicitly controls the allowed distribution deviations.
Specifically, when $T \rightarrow \infty$, the distribution shift  $KL(p(x)\parallel q(x))$ should be small
while for $T\rightarrow 0$, the distribution shift can be very large.
Since $T$ is a pre-defined hyperparameter, $Tlog(N)$ is a constant and can be omitted.
Then, the objective function becomes
\begin{equation}
  F(\omega,W)=\sum_{i=1}^{N} L(x_i,\omega)w_i-T\sum_{i=1}^{N} w_i log w_i -\lambda(\sum_{i=1}^{N} w_i-1). \label{eq:F-lagrange}
\end{equation}
Setting the derivative of $F(\omega,W)$ with respect to $w_i$ to zero,
we obtain the optimality necessary condition for $w_i$, which is
\begin{equation}
w_i={{exp({L(x_i,\omega)\over T})}\over{\sum_{j=1}^{N}{exp({L(x_j,\omega)\over T})}}}. \label{eq:wi-necessary}
\end{equation}
Substituting \eqref{eq:wi-necessary} back into \eqref{eq:F-lagrange},  
we get the effective loss functional
\begin{equation}
F^*(\omega)=T log( {\sum_{i=1}^{N}{exp({L(x_i,\omega)\over T})}}). \label{eq:isloss}
\end{equation}
The minimization of $F^*(\omega)$ with respect to $\omega$ is equivalent to min-max of $R_E(\omega,W)$
with respect to $\omega, W$. 
In this paper, the objective function of the proposed model in \eqref{eq:isloss}
is named as Importance Sampling loss, \textit{ISloss} for short. 

\subsection{LogISloss} \label{sec:log-ISLoss}
In this section, we use the logarithmic transformation $logL(x, \omega)$ instead of $L(x, \omega)$ as a loss measure 
and derive the corresponding objective function. 
The logarithmic risk functional is
\begin{equation}
   R_{log}(\omega) =\int_x log L(x,\omega) p(x) dx.\\
\end{equation}
Similarly, as the derivation for ISloss, the corresponding optimization problem becomes
\begin{align}
  \begin{split} \label{eq:log-problem}
  \min_{\omega} \max_{p(x)} \quad & R_{log}(\omega) =\int_x logL(x,\omega) p(x) dx \\
  \text{s.t.}               \quad & KL(p(x)||q(x)) \leq C. \\
  \end{split}
\end{align}
The optimality necessary condition for $w_{log}^{i}$ is
\begin{equation}
  w_{log}^{i} = \frac{L(x_i, \omega)^{\frac{1}{T}}}{\sum_{j=1}^{N} L(x_i, \omega)^{\frac{1}{T}}} \label{eq:log-wi}
\end{equation}
and the effective empirical loss functional becomes
\begin{equation}
  F^{*}_{log}(\omega) = log (\sum_{i=1}^{N} L(x_i, \omega)^{\frac{1}{T}})^{T}. \label{eq:logisloss}
\end{equation}
In this paper, \eqref{eq:logisloss} is named Logarithmic Importance Sampling loss,
\textbf{LogISloss} for short. 
Equation \eqref{eq:logisloss} shows the minimization problem of 
the logarithmic version $F^*$ is equivalent to the
minimization of the p-norm loss function with $ p={1\over T}$.
The objective function of ISloss\eqref{eq:isloss} involves exponential and not stable in training, 
therefore we use LogISloss\eqref{eq:logisloss} in subsequent experiments. 

\section{Application on face verification task} \label{sec:face-verification}
Face verification\cite{chen2021ring,bianco2017large} is one of the most popular topics 
in the community of computer vision and pattern recognition.
It is a technology used to determine whether a pair of images belong to the same individual and 
is widely used for identity authentication in many areas, such as attendance\cite{pss2016rfid}, 
finance\cite{aru2013facial}, transportation\cite{van2022intelligent}, 
self-service and other fields.
In the face verification community, researchers have proposed large-margin softmax variants
\cite{deng2019arcface,wang2018additive,wang2018cosface,wang2020mis,Liu2022SphereFaceR} 
to enhance the inter-class discrepancy and intra-classs compactness, 
ArcMargin\cite{deng2019arcface} and AddMargin\cite{wang2018additive,wang2018cosface} are two widely-used ones.
In this section, we apply the proposed model to face verification task under ArcMargin and AddMargin.
We denote the cross entropy loss under ArcMargin and AddMargin as \textit{ArcLoss} and \textit{AddLoss}
and derive the logarithmic importance sampling loss under ArcMargin and AddMargin.

\subsection{IS-ArcLoss}
To begin with, let us introduce some notations to facilitate the discussion in this section. 
Suppose $N$ is the number of training samples and $C$ is the number of classes.
$x_i \in \mathcal{R}^d$ denotes the deep feature of $i$-th sample.
For $x_i$, we refer $y_i$ as the target class and $j$ as the non-target class (the label excluding the ground truth one).
$W_j \in \mathcal{R}^d$ denotes the $j$-th column of the weight $W \in \mathcal{R}^{d\times C}$
and $b_j \in \mathcal{R}^C$ is the bias term. We fix bias $b_j=0$ for simplicity.
Suppose $\theta_{j,i}$ is the angle between $W_j$ and $x_i$. 
Following \cite{deng2019arcface, wang2018cosface}, we fix the weight
$\norm{W_j}=1$ by $l_2$ normalization and fix the deep feature 
$\norm{x_i}$ by $l_2$ normalization and rescale it to $s$.
The feature scale $s$ is set to 64 following \cite{deng2019arcface,wang2018cosface}.

ArcFace(Additive Angular Margin Loss)\cite{deng2019arcface}
uses the arc-cosine function to calculate the angle between the feature $x_i$ and the target weight $W_{y_i}$.
The model adds an additive angular margin $m$ to the \textit{target angle} 
to enhance the inter-class discrepancy and intra-class compactness. 
The ArcLoss for the $i$-th sample is defined as follows
\begin{equation}
  L_{i}^{Arc} = - log \frac{e^{s(cos(\theta_{y_i,i} + m))}}
             {e^{s(cos(\theta_{y_i,i} + m))}+\sum_{j=1,j\neq y_i}^{C} e^{s cos(\theta_{j,i})}}. \label{eq:arc-margin}
\end{equation}
Suppose $L_{i}^{Arc}$ denote the ArcLoss for the $i$-th sample,
and $n$ denote the number of samples in each training batch,
then we name the proposed logarithmic importance sampling loss as \textit{IS-ArcLoss}, 
which is defined as follows
\begin{equation}
  L^{IS}_{Arc} = log (\sum_{i=1}^{n}(L_{i}^{Arc})^{1\over T})^{T}. \label{eq:IS-ArcMargin}
\end{equation}
Here, the superscript \textit{IS} denotes \textbf{I}mportance \textbf{S}ampling. 
In subsequent experiments for ArcLoss and IS-ArcLoss, the angular margin penalty $m$
is fixed to 0.5 as recommended by \cite{deng2019arcface}.

\subsection{IS-AddLoss} 
AddMargin(Additive margin)\cite{wang2018additive}, 
also known as the Large-Margin Cosine Loss\cite{wang2018cosface},
calculates the angle between the feature $x_i$ and the target weight $W_{y_i}$.
The model minus an additive margin $m$ to the cosine of the target angle
to maximize inter-class variance and minimize intra-class variance. 
The AddLoss for the $i$-th sample is defined as follows
\begin{equation}
  L_{i}^{Add} =  - log \frac{e^{s(cos(\theta_{y_i,i})-m)}}
            {e^{s(cos(\theta_{y_i,i})-m)} + \sum_{j=1,j\neq y_i}^{C}e^{s cos(\theta_{j,i})}}. \label{eq:add-margin}
\end{equation}
Suppose $L_{i}^{Add}$ denote the AddLoss for the $i$-th sample,
then we name the proposed logarithmic importance sampling loss as \textit{IS-AddLoss}, 
which is defined as follows
\begin{equation}
  L^{IS}_{Add} = log (\{\sum_{i=1}^{n}(L_{i}^{Add})^{1\over T}\})^{T}. \label{eq:IS-AddMargin}
\end{equation}
In subsequent experiments for AddLoss and IS-AddLoss, the additive margin $m$ 
is fixed to 0.35 as recommended by \cite{wang2018cosface}.

\section{Experiments and Results} \label{sec:experiments}
In this section, we conducted experiments to show the effectiveness of the proposed method. 
\Cref{sec:datasets} introduces datasets including training datasets, test datasets and the training schedule. 
\Cref{sec:weights} analyzes the importance sampling weights. 
\Cref{sec:temperature} describes how the hyperparameter $T$ affects the model performance. 
\Cref{sec:distributionDeviation} compares ArcLoss and IS-ArcLoss, AddLoss and IS-AddLoss with respect to distribution deviations.
\Cref{sec:Hard Pairs} compares ArcLoss and IS-ArcLoss with respect to hard examples.  

\subsection{Datasets} \label{sec:datasets}
In the face verification system, training stage and test stage are different.  
The training stage is used to learn the deep feature embedding for a given image through a classification task.  
In the test stage, the score of a given image pair is usually calculated 
by the cosine similarity between the two feature embeddings.
If the score is higher than a given threshold, the input pair is considered a positive pair.
If the score is lower than a given threshold, the input pair is considered a negative pair.
The given verification threshold is obtained from cross-validation by a pre-defined split in test datasets. 

\textbf{Training Datasets}
In this paper, we uses the Racial Faces in-the-Wild(RFW) database\cite{Wang_2019_ICCV}, 
which includes four races, namely Caucasian, African, Asian and Indian.
We use BUPT-Equalizedface\cite{Wang_2019_ICCV} as the training dataset which 
contains 27999 individuals and 1251416 images.
The number of individuals in Caucasian, African, Asian and Indian
are 7000, 7000, 7000 and 6999 respectively.
The detailed statistics of BUPT-Equalizedface are given in Appendix-A. 
In later discussions, the four training subsets in BUPT-Equalizedface are denoted as 
RFW-Caucasian, RFW-African, RFW-Asian and RFW-Indian.

\begin{figure}
  \begin{center}
  \includegraphics[width=\linewidth]{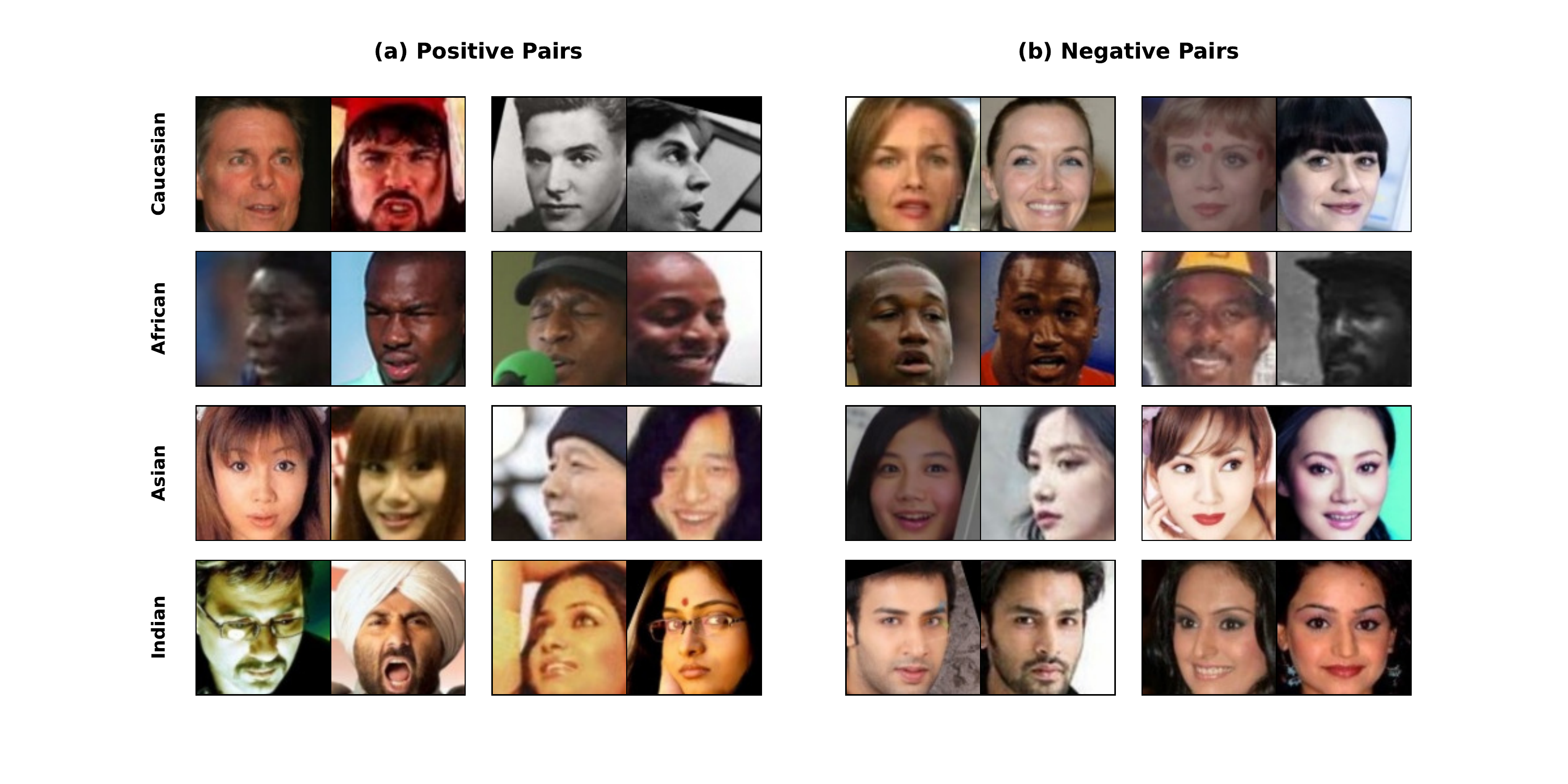}
  \end{center}
\caption{Images from RFW-test. 
Four rows show examples from Caucasian, African, Asian and Indian respectively. 
(a) shows hard positive pairs and (b) shows hard negative pairs.}
\label{fig:RFW-imgs}
\end{figure}

\textbf{Test Datasets}
The RFW database has four test subsets, namely Caucasian, African, Asian and Indian.
There are about 14K positive pairs and 50M negative pairs\cite{Wang_2019_ICCV} 
for each subset in the RFW-test, most of which are easy to distinguish.  
In this work, the proposed model aims to emphasize large distribution deviations and hard samples. 
Therefore, we use the RFW-test for evaluation, which is composed of 
difficult pairs selected based on cosine similarity between embeddings of each pair\cite{Wang_2019_ICCV}. 
Each subset (race) contains 3000 positive pairs and 3000 negative pairs, splited into 10 subsets in advance.
We selected the verification threshold for one split based on the remaining nine splits. 
We propose RFW-extend as an extension to RFW-test and the details of the dataset are shown in \Cref{sec:Hard Pairs}.
\autoref{fig:RFW-imgs} shows selected positive pairs and negative pairs from each race.
It can be seen that some pairs are difficult even for human observers. 

\textbf{Experiment Settings}
The datasets were aligned and preprocessed in MXNet binary format
and the size of the cropped image were set to $112 \times 112$.
We set the training batch size, weight decay and momentum as 128, 5e-4 and 0.9, respectively.
The dimension of embedding feature was set to 512
following \cite{deng2019arcface,wang2018cosface}.  
The initial learning rate was set to 0.1, and decreased by a factor of 10 at given epochs.
For the BUPT-Equalizedface dataset, we trained the network for 27 epochs 
and decayed the learning rate at epochs 14, 20 and 24.
We used half of the suggested epochs in \cite{MetaRFW} to save training time. 
We used IResNet-34\cite{deng2019arcface}, a 34-layer modified ResNet,
as the default backbone in all experiments if not otherwise specified.
We followed the Pytorch implementation of 
insightface\footnote{\url{https://github.com/deepinsight/insightface/tree/master/recognition/arcface_torch}} 
on one GTX 1080ti and one GTX 3090ti for training. 

\subsection{Importance sampling weights} \label{sec:weights}
\begin{figure}
  \begin{center}
  \includegraphics[width=.8\linewidth]{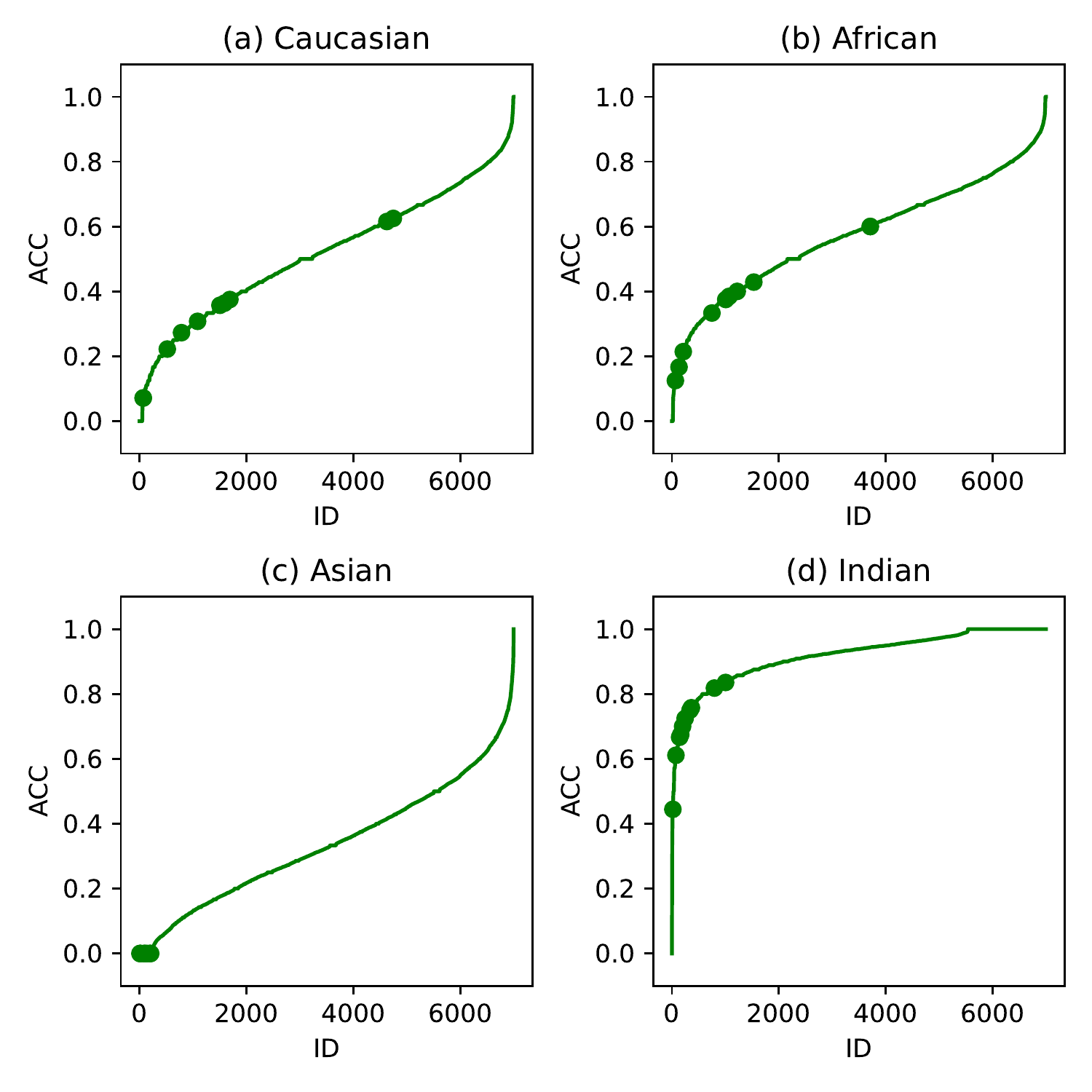}
  \end{center}
\caption{Classification accuracy and the top-10 importance sampling weights. 
X-axis represents the individual (ID) index.
Y-axis represents the average accuracy for each ID (ascending order).
The top-10 maximum importance sampling weights are marked in green points.}
\label{fig:weight}
\end{figure}

In this section, we give a heuristic example to show that
the importance sampling weights emphasize the hard samples in the training dataset. 
We calculated the average classification accuracy for each individual (ID) 
and highlighted the individuals with the top-10 maximum average weights.
We first calculated the cross entropy loss for each training sample 
then used \eqref{eq:log-wi} to calculate the weight.
All models in \autoref{fig:weight} were trained under IS-ArcLoss($T=0.5$).
\autoref{fig:weight} shows that the individuals with maximum weights 
are those with low accuracy,  
especially in (c) where the top-10 maximum weights clustered around 0.
Hard samples are normally assumed to have low classification accuracy.
Therefore, this observation showed the importance sampling weight 
emphasized hard training samples.
We further analyze how the importance sampling weights change over time in Appendix-B.
The experiment results indicated the importance sampling weights 
gradually concentrated on hard samples as the training proceeded. 

\subsection{Temperature} \label{sec:temperature}
\begin{figure}
  \begin{center}
  \includegraphics[width=.8\linewidth]{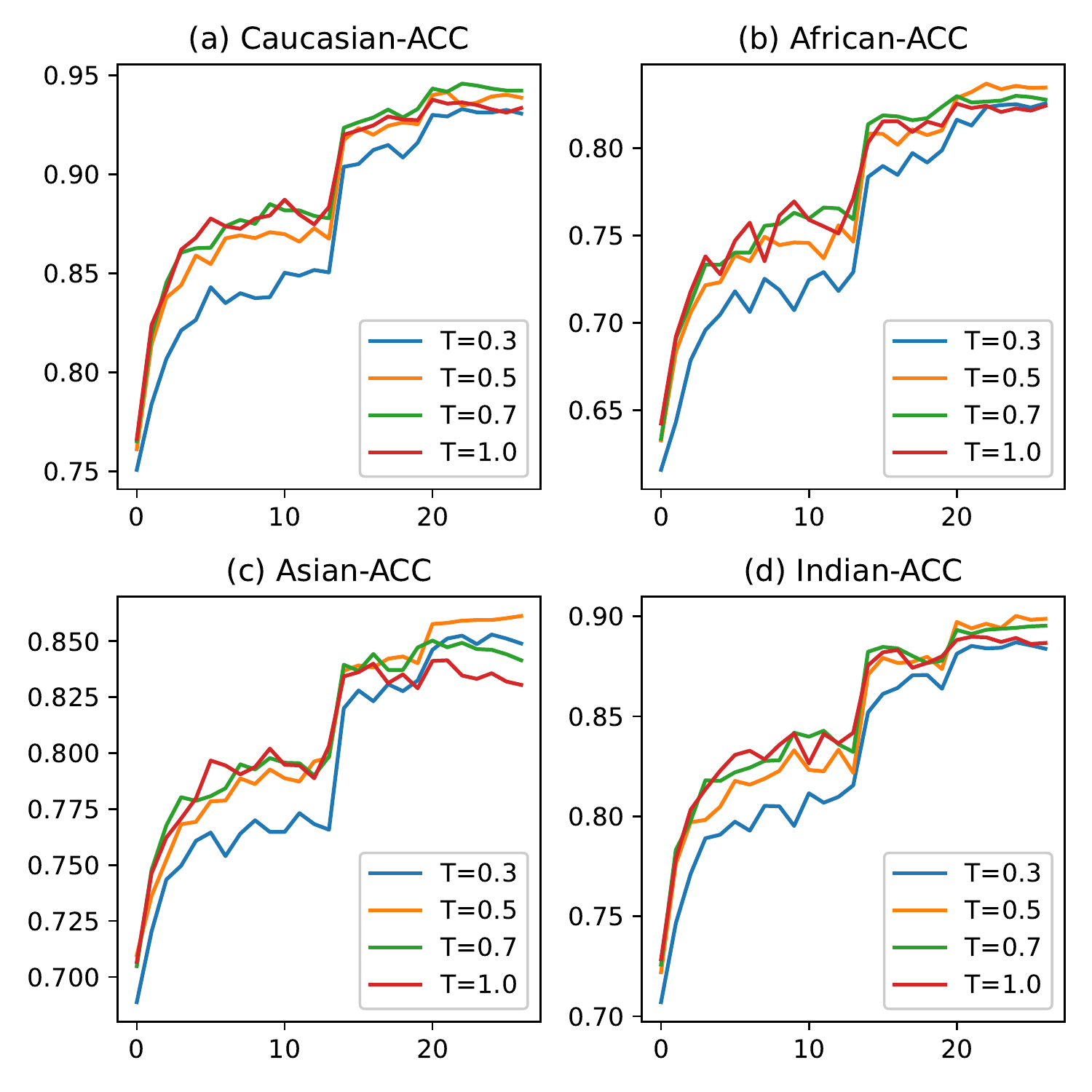}
  \end{center}
\caption{Comparison of verification accuracy among different $T$.
X-axis represents training epochs while Y-axis represents verification accuracy.
The model was trained on RFW-Caucasian using IS-ArcLoss.}
\label{fig:diffT}
\end{figure}

In this section, we analyze how the temperature $T$ affects the model performance.
\autoref{fig:diffT} compares verification accuracy on four RFW-test datasets when $T=0.3,0.5,0.7,1.0$.
\autoref{fig:diffT} shows that $T=0.5$ achieves the best accuracy on African, Asian and Indian 
while $T=0.7$ achieves the best accuracy on Caucasian. 
One possible explanation is that the distribution deviation between RFW-Caucasian (training) and Caucasian(test)
is smaller than the distribution deviations between RFW-Caucasian (training) and African/Asian/Indian(test), 
therefore, the test Caucasian favors a higher temperature, $T=0.7$.
This observation suggests that lower $T$ tolerates larger distribution deviations. 
In this paper, we used a fixed $T=0.5$ as the default temperature in all subsequent experiments.
However, other training schedules on $T$ can be adopted.
For example, starting $T$ at a high temperature and decaying it during training.
How to choose a good schedule on $T$ for different learning problems will be explored in our future studies. 

\subsection{Distribution deviations} \label{sec:distributionDeviation}
\begin{table}
  \caption{\label{tab:race-arcmargin}
   Verification accuracy(\%) of IResNet-34 models trained on four datasets.
   ``R-Cau, R-Afr, R-Asi, R-Ind'' represents RFW-Caucasian, RFW-African, RFW-Asian and RFW-Indian respectively.
   \textit{Arc} represents ArcLoss and \textit{IS-Arc} represents IS-ArcLoss.
   Results are in \% and a higher number indicates better performance.}
   \centering
  \begin{tabular}{ llllll }
      \hline
      Train & Loss & Cau & Afr & Asi & Ind  \\
      \hline
      \multirow{ 2}{*}{R-Cau}   &Arc    & 93.55 & 82.23 & 84.70 & 88.77 \\
                                &IS-Arc & 94.15\tiny{(0.60)} & 83.22\tiny{(0.99)} & 85.82\tiny{(1.12)} & 89.40\tiny{(0.63)}\\
      \hline
      \multirow{ 2}{*}{R-Afr}   &Arc    & 89.23 & 92.37 & 84.02 & 88.20\\
                                &IS-Arc & 90.37\tiny{(1.14)} & 92.43\tiny{(0.06)} & 85.82\tiny{(1.80)} & 89.65\tiny{(1.45)}\\
      \hline
      \multirow{ 2}{*}{R-Asi}   &Arc    & 85.50 & 79.13 & 90.90 & 84.45\\
                                &IS-Arc & 87.20\tiny{(1.70)} & 80.30\tiny{(1.17)} & 91.17\tiny{(0.27)} & 85.32\tiny{(0.87)}\\
      \hline
      \multirow{ 2}{*}{R-Ind}   &Arc   & 87.98 & 81.47 & 83.60 & 91.50\\
                                &IS-Arc & 88.92\tiny{(0.94)} & 82.30\tiny{(0.83)} & 83.48\tiny{(-0.12)} & 91.80\tiny{(0.3)}\\
      \hline 
  \end{tabular}
\end{table}
In this section, we compare the performance of LogISLoss and cross-entropy loss 
with respect to distribution deviations. 
\autoref{tab:race-arcmargin} compares verification accuracy between ArcLoss and IS-ArcLoss.
The models were trained on RFW-Caucasian, RFW-African, RFW-Asian, RFW-Indian
and tested on Caucasian, African, Asian, Indian.
We report the result at the epoch where the training race achieves the highest verification accuracy.
The accuracy gain (in small brackets) is the difference between our proposed IS-ArcLoss over ArcLoss. 
For example, the accuracy of the model trained on RFW-Caucasian and tested on African is 82.23\% 
of ArcLoss and 83.22\% of IS-ArcLoss, the accuracy gain is 0.99\%.
\autoref{tab:race-arcmargin} shows the following findings.
First, IS-ArcLoss performs better than ArcLoss in most cases.
Second, the accuracy gains in non-training races are higher compared with the training race
on three test datasets(RFW-Caucasian, RFW-African, RFW-Asian).
Take the models trained on RFW-Caucasian(first row) for example, the accuracy gain on 
Caucasian, African, Asian, Indian is 0.6\%, 0.99\%, 1.12\% and 0.63\% respectively.
The accuracy gain on Caucasian(0.6\%) is the lowest. 
This observation indicates that IS-ArcLoss has better generalization ability compared with ArcLoss. 
If we assume the accuracy gap is a measure of the distribution shift,
then a larger accuracy gap indicates a larger distribution shift.  
Next, consider the model trained on RFW-African(second row) for example, 
the test accuracy on Asian, Indian and Caucasian is 84.02\%, 88.20\% and 89.23\%.    
Under the previous assumption, the distribution deviations between
Asian, Indian, Caucasian and African are in increasing order.
The accuracy gain on the three datasets is 1.80\%, 1.45\% and 1.14\%, which is in decreasing order. 
This observation implies that the proposed IS-ArcLoss performs better 
when the distribution deviation is large. 
More experiment results on IS-ArcLoss with respect to TAR@FAR
and on different backbones are shown in Appendix-C.

\begin{table}
  \caption{\label{tab:race-addmargin}
   Verification accuracy(\%) of IResNet-34 models trained on different training datasets.
   \textit{Add} represents AddLoss and \textit{IS-Add} represents IS-AddLoss.}
   \centering
  \begin{tabular}{ llllll }
      \hline
      Train & Loss & Cau & Afr & Asi & Ind  \\
      \hline
      \multirow{ 2}{*}{R-Cau}   &Add    & 94.28 & 83.27 & 85.70 & 89.45\\
                                &IS-Add & 93.68\tiny{(-0.60)} & 83.07\tiny{(-0.20)} & 85.97\tiny{(0.27)} & 89.68\tiny{(0.23)}\\
      \hline
      \multirow{ 2}{*}{R-Afr}   &Add    & 90.85 & 92.72 & 84.85 & 89.85\\
                                &IS-Add & 89.98\tiny{(-0.87)} & 92.50\tiny{(-0.22)} & 86.10\tiny{(1.25)} & 89.93\tiny{(0.08)}\\
      \hline
      \multirow{ 2}{*}{R-Asi}   &Add    & 86.75 & 79.62 & 91.17 & 85.22\\
                                &IS-Add & 86.87\tiny{(0.12)} & 80.63\tiny{(1.01)} & 90.28\tiny{(-0.89)} & 85.75\tiny{(0.53)}\\
      \hline
      \multirow{ 2}{*}{R-Ind}   &Add    & 88.93 & 82.78 & 83.93 & 92.13\\
                                &IS-Add & 88.47\tiny{(-0.46)} & 82.12\tiny{(-0.66)} & 83.02\tiny{(-0.91)} & 91.20\tiny{(-0.93)}\\
      \hline 
  \end{tabular}
\end{table}
\autoref{tab:race-addmargin} compares verification accuracy between AddLoss and IS-AddLoss.
IS-AddLoss performs better than AddLoss on 3(RFW-Caucasian, RFW-African, RFW-Asian) out of 4 test datasets 
on non-training races. 
Another interesting observation is in line with the observation in \autoref{tab:race-arcmargin}.
For the models trained on RFW-African, the accuracy gain on Asian(84.85\%), Indian(89.85\%) and Caucasian(90.85\%)
is 1.25\%, 0.08\% and -0.87\%, which is in decreasing order. 
For the models trained on RFW-Asian, the accuracy gain on African(79.62\%), Indian(85.22\%) and Caucasian(86.75\%) 
is 1.01\%, 0.53\% and 0.12\%, which is in decreasing order. 
The results also confirm our findings in \autoref{tab:race-arcmargin} that LogISloss performs better 
when the distribution shift is large. 

\autoref{tab:other-arcmargin} test LogISloss on several popular benchmarks, 
including Labeled Faces in the Wild(LFW)\cite{huang2008labeled},
Age-DB\cite{moschoglou2017agedb},
Cross-age LFW(CALFW)\cite{zheng2017cross},
Cross-pose LFW(CPLFW)\cite{zheng2018cross} and 
Celebrities in Frontal-Profile  (CFP-FP and CFP-FF)\cite{sengupta2016frontal}, 
where LFW, AGE-DB, CALFW and CPLFW have 6000 pairs and CFP(large pose variation) has 7000 pairs.
These datasets are also split into 10 subsets in advance.
The results were evaluated on the final epoch. 
\autoref{tab:other-arcmargin} shows that IS-ArcLoss performs better than ArcLoss in most cases.
To be specific, the accuracy gains on LFW and CFP-FF are low, 
the accuracy gains on CALFW and CPLFW are medium, 
and the accuracy gains on CFP-FP and AgeDB are high. 

\begin{table*} 
  \caption{\label{tab:other-arcmargin}
    Verification accuracy (\%) of IResNet-34 models tested on different datasets.}
   \centering
  \begin{tabular}{ llllllll p{8cm} }
      \hline
      Train & Loss & LFW & CFP-FP & CFP-FF & AgeDB & CALFW & CPLFW  \\
      \hline
      \multirow{ 2}{*}{RFW-Caucasian} &ArcLoss    & 99.30 & 86.39 & 98.90 & 94.67 & 92.28 & 84.40 \\
                                      &IS-ArcLoss & 99.33(0.03) & 88.41(2.02) & 99.11(0.21) & 95.47(0.80)& 92.87(0.59) & 84.90(0.50) \\
      \hline
      \multirow{ 2}{*}{RFW-African} &ArcLoss     & 99.12 & 86.03 & 98.77 & 90.68 & 91.37 & 82.82 \\
                                    &IS-ArcLoss  & 99.10(-0.02) & 87.09(1.06) & 98.97(0.20) & 92.20(1.52)& 91.77(0.40) & 83.62(0.80) \\
      \hline
      \multirow{ 2}{*}{RFW-Asian}   &ArcLoss     & 98.13 & 89.46 & 97.49 & 89.55 & 88.42 & 82.65 \\
                                    &IS-ArcLoss  & 98.62(0.49) & 91.51(2.05) & 97.37(-0.12) & 90.30(0.75)& 89.28(0.86) & 83.17(0.52) \\
      \hline
      \multirow{ 2}{*}{RFW-Indian}  &ArcLoss     & 98.62 & 92.30 & 98.27 & 89.28 & 89.32 & 84.93 \\
                                    &IS-ArcLoss  & 98.68(0.06) & 92.53(0.23) & 98.41(0.14) & 90.77(1.49)& 89.42(0.10) & 85.57(0.64) \\
      \hline
  \end{tabular}
\end{table*}

\subsection{Hard Pairs} \label{sec:Hard Pairs}
\begin{figure}
  \begin{center}
  \includegraphics[width=\linewidth]{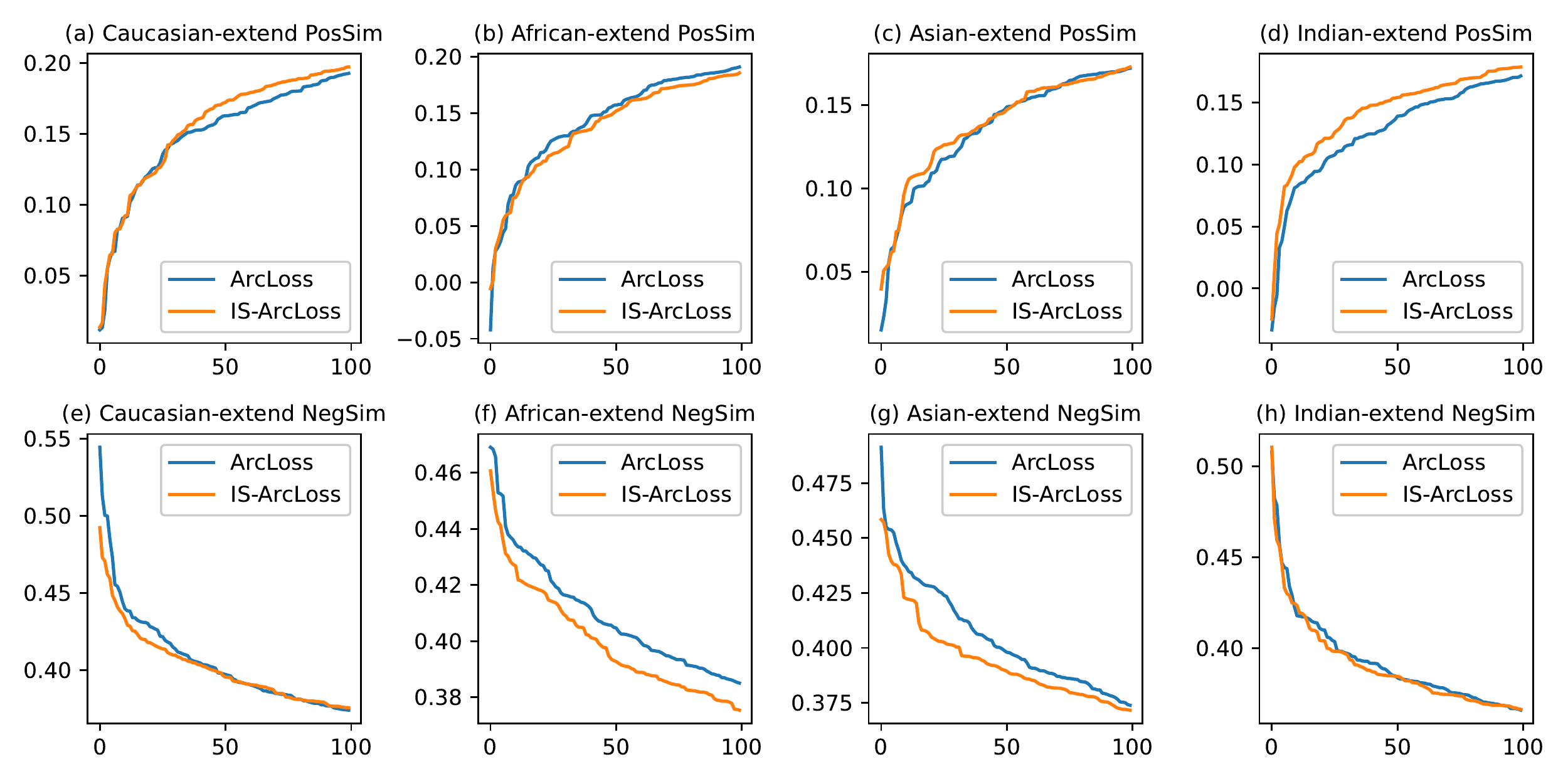}
  \end{center}
\caption{Hard pairs' similarity comparison between ArcLoss and IS-ArcLoss.  
The model was trained on RFW-Equalizedface and tested on the RFW-test-extend dataset.
(a) (b) (c) (d) show positive similarity while (e) (f) (g) (h) show negative similarity.}
\label{fig:plotWorst}
\end{figure}

\textbf{RFW-test-extend Dataset}
In this experiment, we propose an extended test dataset \textit{RFW-test-extend} to include more pairs for evaluation. 
The extended dataset has four subsets, namely Caucasian-extend, African-extend, Asian-extend and Indian-extend.
The number of positive pairs for the four subsets are 13650,14050,15690,13860
and the number of negative pairs are 42260,43370,54890,43120.
All positive pairs and negative pairs are randomly split into 10 folds. 
The generation and statistics for RFW-extend are shown in Appendix-A.
In face verification task, for positive pairs, the similarity between two images should be high, 
therefore \textbf{hard positive} pairs have low similarity. 
For negative pairs, the similarity between two images should be low, 
therefore \textbf{hard negative} pairs have high similarity.
\autoref{fig:plotWorst} shows the similarity of top-100 hard positive and hard negative pairs.  
The models were trained on the whole BUPT-Equalizedface datasets (27999 IDs) under ArcMargin.
\autoref{fig:plotWorst} shows that IS-ArcLoss has higher similarity for positive pairs 
compared with ArcLoss in (a)(Caucasian-extend), (c)(Asian-extend), (d)(Indian-extend)
and it has lower similarity for negative pairs in all tested datasets.
Visualization of the top-2 hardest pairs for each race is shown in \autoref{fig:RFW-imgs}.
This observation indicates that IS-ArcLoss emphasizes the hardest pairs. 
\section{\textbf{Conclusion}} \label{sec:conclusion}
In this paper, we develop an information-theoretic learning model derived from importance sampling to deal with 
the problem of data distribution deviation between current training data and future test data. 
We reveal that minimization of the objective function of LogISloss 
is equivalent to the minimization of the p-norm loss function when $p=\frac{1}{T}$.
We applied the proposed model to the face verification task.
Experiments on distribution deviations showed that the learned features under LogISloss generalized well
across different races and across different datasets compared with cross-entropy loss and 
experiments on the hardest pairs showed that LogISloss emphasized the hard pairs. 
Applying ISloss/LogISloss to other machine learning problems 
and designing a good schedule for $T$ remains to be explored in future endeavors. 
\section*{\textbf{Disclosure statement}} 
No potential conflict of interest was reported by the authors.
\section*{Acknowledgements} 
This work is supported by the National Natural Science Foundation of China under Grants 61976174.
\bibliographystyle{elsarticle-num} 
\bibliography{ref} 

\end{document}